%% file: main.tex
\pdfoutput=1

\documentclass[final]{cvpr}

\usepackage{times}
\usepackage{epsfig}
\usepackage{graphicx}
\usepackage{amsmath}
\usepackage{amssymb}
\usepackage{enumitem}

\input{math_commands.tex}

\usepackage[pagebackref=true,breaklinks=true,colorlinks,bookmarks=false]{hyperref}

\usepackage[ruled,vlined,linesnumbered]{algorithm2e}
\usepackage{subcaption}
\usepackage{threeparttable}
\usepackage{booktabs}

\newif\ifmain
\maintrue

\newif\ifappendix
\appendixtrue

\def\cvprPaperID{6464} 
\def\confYear{CVPR 2021}

\begin{document}

\ifmain

\title{Efficient Model Performance Estimation via Feature Histories}

\author{Shengcao Cao \quad Xiaofang Wang \quad Kris Kitani\\
Carnegie Mellon University\\
{\tt\small \{shengcao,xiaofan2,kkitani\}@cs.cmu.edu}
}

\maketitle

\begin{abstract}
An important step in the task of neural network design, such as hyper-parameter optimization (HPO) or neural architecture search (NAS), is the evaluation of a candidate model's performance. Given fixed computational resources, one can either invest more time training each model to obtain more accurate estimates of final performance, or spend more time exploring a greater variety of models in the configuration space. In this work, we aim to optimize this exploration-exploitation trade-off in the context of HPO and NAS for image classification by accurately approximating a model's maximal performance early in the training process. In contrast to recent accelerated NAS methods customized for certain search spaces, \eg, requiring the search space to be differentiable, our method is flexible and imposes almost no constraints on the search space. Our method uses the evolution history of features of a network during the early stages of training to build a proxy classifier that matches the peak performance of the network under consideration. We show that our method can be combined with multiple search algorithms to find better solutions to a wide range of tasks in HPO and NAS. Using a sampling-based search algorithm and parallel computing, our method can find an architecture which is better than DARTS and with an 80\% reduction in wall-clock search time.

\end{abstract}

\section{Introduction}

Identifying the optimal hyperparameters or the best architecture is important for maximizing the performance of neural networks. Accordingly, algorithms for hyperparameter optimization (HPO) and neural architecture search (NAS) have been proposed to automatically select the optimal hyperparameters and architectures in a data-driven manner. Existing HPO or NAS methods typically require that many candidate configurations of hyperparameters or architectures are evaluated. However, such evaluation is extremely expensive as fully training one model until convergence may take several GPU days given the complexity of modern deep learning models and datasets. This calls for efficient methods that can approximate the model's final performance early in the training process.

\begin{figure}[t]
\begin{center}
\includegraphics[width=1.0\linewidth]{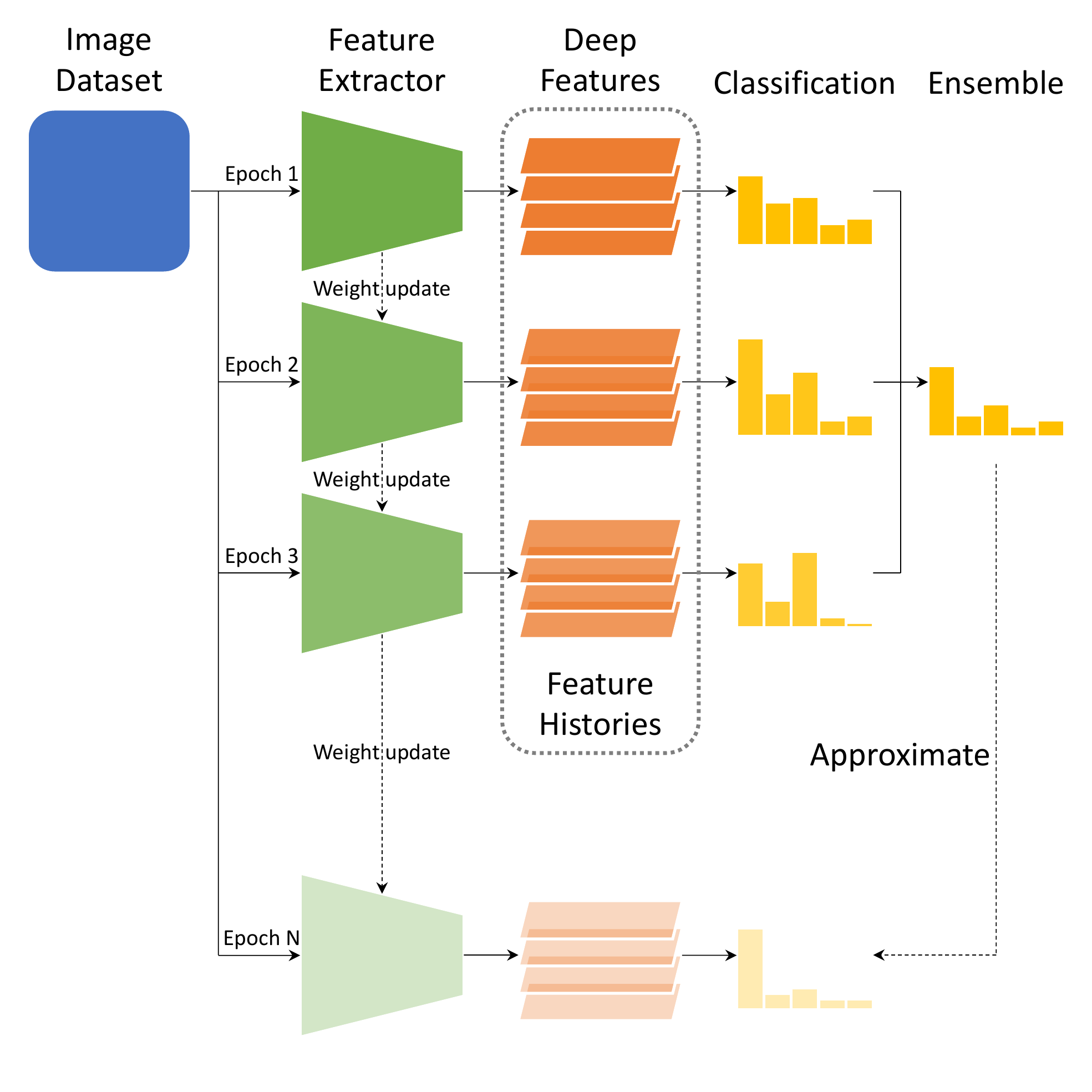}
\caption{An illustration of our proposed proxy classifier using feature histories. Our method aims to approximate the model's final performance without actually training it until convergence. The major steps are: 1) collecting feature histories of early training stages, 2) finding the optimal linear classifiers for each checkpoint, and 3) ensembling them to build a proxy classifier that can simulate the outputs of the model at convergence.}
\label{fig-main}
\end{center}
\end{figure}

We propose to leverage \textit{feature histories}, that is, checkpoints of evolving deep features prior to the final fully-connected layer of a neural network during training, to predict what the output of the network might be at convergence. Our proposed approach is motivated by this key observation: Many images which are correctly classified when the network converges, keep changing between being classified correctly and incorrectly at the early stage of training. Only using a particular checkpoint early in the training will lead to an inaccurate approximation for the final accuracy. However, for a particular image, the consensus of classification results by early checkpoints is usually the same as the final model's decision. Therefore, we propose to build a proxy classifier based on the early feature histories, whose behavior coincides with the network at convergence, and use its evaluated performance as an approximation of the model's final performance. In other words, to approximate a candidate model's final performance, we no longer need to train it until convergence, but create a proxy based on the feature histories to simulate the converged network. This proxy classifier can bring better assessment of candidate models within limited training budgets, thus benefiting the task of HPO and NAS. A brief illustration of our method is shown in Figure~\ref{fig-main}. We will support this observation with experimental results in Section~\ref{sec-method-observation}. Our empirical results show that this method applies to various architecture families, including VGG~\cite{simonyan2014very}, ResNet~\cite{he2016deep}, MobileNet~\cite{sandler2018mobilenetv2}, and also architectures in different NAS tasks, including DARTS~\cite{liu2018darts} and NASBench-201~\cite{dong2020bench}.


Our proposed method has these advantages:

\begin{itemize}[leftmargin=*]
    \vspace{-0.5em}
    \setlength\itemsep{0.01em}
    \item \textbf{Versatility}: Our method does not depend on any pre-training, external data, or specific search space, but only makes use of the feature histories. It can be incorporated into several search algorithms for general purpose, and applied to a wide range of tasks in HPO and NAS.
    \item \textbf{Accuracy}: Experimental results show that the performance of our proposed proxy classifier can accurately approximate the final performance of given candidate model configurations. The relative performance ranking of configurations is also well preserved.
    \item \textbf{Efficiency}: Our method is able to approximate the final performance using only partially trained models. The extra overhead is no more than one epoch of training time.
\end{itemize}

\section{Related Work}
We propose a novel method for approximating a model's final performance at an early training stage, which can be applied to hyper-parameter optimization (HPO) and neural architecture search (NAS) tasks. Here we briefly summarize the previous work that our method is based on.

\paragraph{Performance estimation}
To accelerate hyperparameter optimization, \cite{domhan2015speeding} proposes to extrapolate the learning curves based on early training stage and terminate bad configurations accordingly. \cite{klein2016learning} improves the estimation of the learning curve with a Bayesian neural network. \cite{baker2017accelerating} applies this strategy to NAS, using architectural hyperparameters for the learning curve prediction. In contrast to these approaches, our method use feature histories during network optimization, instead of performance metrics, to approximate the final performance. Our method does not require training another prediction network, or manually designed learning curve modeling. It has also been shown that a carefully selected learning rate schedule can produce high-performance models with a limited budget~\cite{li2019budgeted}. However, this type of learning rate schedule is impractical in the case where we need to dynamically adjust the budget, which is common in hyperparameter optimization algorithms. Other methods specifically designed for accelerating performance estimation in NAS includes inheriting network weights through network morphisms~\cite{chen2015net2net}, weight sharing across architectures~\cite{bender2018oneshot, pham2018enas, shi2020bridging}, and optimize budgeted performance estimation~\cite{zheng2020rethinking}. These methods are usually constrained by the customized architecture search space, while our method can be easily extended to various tasks.

\paragraph{Hyperparameter optimization}
Automated hyperparameter optimization can improve the performance of deep learning models while reducing human effort. Model-free methods including random search and grid search, can be considered as the most basic HPO approach. Bayesian optimization, an effective optimization method for computational costly functions, has several variants~\cite{bergstra2011algorithms, snoek2012practical, snoek2015scalable}. More recently, \cite{li2017hyperband} proposes a bandit-based strategy named HyperBand, which dynamically allocates a budget for configurations based on evaluation results. BOHB~\cite{falkner2018bohb} combines Bayesian optimization and HyperBand to achieve the best of both worlds. Our experimental validation builds on HyperBand and BOHB and shows how our method can further improve hyperparameter optimization.

\paragraph{Neural architecture search}
Automated architecture design can discover architectures that outperform manually designed CNNs. For a comprehensive overview of NAS, one may refer to \cite{elsken2018neural}. The search methods include reinforcement learning~\cite{zoph2016neural, zoph2018learning, tan2019mnasnet}, evolutionary algorithm~\cite{real2017large, xie2017genetic, real2019regularized}, \etc. Gradient-based NAS methods~\cite{liu2018darts, chen2019progressive, xu2019pc} greatly reduce the search cost by relaxing the discrete search space to be continuous and applying gradient descent. For better evaluation and comparison of different NAS algorithms, some benchmarks~\cite{ying2019bench, klein2019tabular, dong2020bench} have been proposed.


\paragraph{Ensemble of history models}
\cite{xie2013horizontal} proposes to combine the outputs of networks from earlier epochs to improve final predictions. \cite{huang2017snapshot} uses cyclic learning rate to efficiently obtain multiple well-trained models and their ensemble. \cite{garipov2018loss} proposes a method to discover high-accuracy paths between trained models and their ensemble. Our method can be considered as an ensemble of history models as well. In contrast to previous methods, we focus on ensembling simple linear classifiers based on feature histories, which greatly improves the efficiency of model performance evaluation.

\section{Approach}

We first describe the observation that motivates this work and the details of our proposed method for efficient performance estimation in Section~\ref{sec-method-main}, and then explain how to combine our method with other search algorithms for neural architecture search (NAS) and hyperparameter optimization (HPO) applications in Section~\ref{sec-method-comb}.

\subsection{Performance Estimation via Feature Histories}
\label{sec-method-main}

\subsubsection{Preliminaries}

A modern CNN designed for image classification~\cite{krizhevsky2012imagenet, simonyan2014very, he2016deep, xie2017aggregated} can be divided into two parts: the feature extractor and the linear classifier. The feature extractor part typically consists of multiple convolutional layers, normalization layers, and pooling layers. The learned feature mapping can often be transferred to other datasets or tasks including object detection~\cite{ren2015faster} and segmentation~\cite{he2017mask}. In contrast, the linear classifier is one single fully connected layer, which outputs logits for each image class and learns the optimal linear boundaries with respect to the loss function. Compared with the linear classifier, the feature extractor has more parameters, is more computationally expensive and harder to optimize. Formally, we can consider a CNN as the composition of two functions:
$$f(\vx;\vw^{fea}, \vw^{cls})=h(g(\vx; \vw^{fea});\vw^{cls})$$
where $f:\sX\to\R^c$ is a CNN which maps the input image space $\sX$ to $c$ class logits, $g:\sX\to\R^d$ is the feature extractor with parameters $\vw^{fea}$ which maps images to a $d$-dim deep feature space, and $h:\R^d\to\R^c$ is the linear classifier with parameters $\vw^{cls}$ which maps deep features to $c$ class logits. The task of the feature extractor is to learn a mapping from images to deep features such that the features corresponding to different image classes are linearly separable. Given a fixed feature extractor mapping, it is easy to find the optimal linear boundaries in the feature space.

\subsubsection{Observation about Feature Histories}
\label{sec-method-observation}

During the process of CNN optimization and evaluation, we observe these phenomena:
\begin{itemize}[leftmargin=*]
    \vspace{-0.5em}
    \setlength\itemsep{0.01em}
    \item \textbf{Classification oscillation}: During training, the features of most images fluctuates between being classified correctly and incorrectly by the optimal linear classifier. As training continues, the classification results will become stabilized and more features will lie on the correct side of the decision boundary.
    \item \textbf{Consistency between historical models}: If the feature of one validation image lies on the correct side of the optimal linear boundaries most of the time, with high probability it will be correctly classified at the end of optimization. Similarly, if the feature is mostly on the wrong side of the decision boundary in history, it will be more likely to be mis-classified by the model at convergence.
\end{itemize}

Next we will use an example of training ResNet-18 on CIFAR-100 to demonstrate this observation qualitatively and quantitatively. In fact, this observation generalizes well to various CNN architectures, as we will show in Section~\ref{sec-exp}.
We train ResNet-18 for $200$ epochs and save the deep features of training and validation images collected in every epoch, \ie, feature histories throughout training. We compare three classifiers:
\begin{itemize}[leftmargin=*]
    \vspace{-0.5em}
    \setlength\itemsep{0.01em}
    \item \textbf{Original}: The saved linear classifier at the given epoch. Evaluating this classifier with corresponding saved deep features is identical to evaluating the original ResNet-18 checkpoints.
    \item \textbf{Optimized}: We can also optimize the linear classifiers based on the saved training image features at given epochs and ground truth labels. This classifier represents the optimal linear boundaries of feature histories. Note that separately optimizing the linear classifiers with saved features requires much less computational resources compared with training the whole ResNet-18.
    \item \textbf{Ensemble}: Based on the optimal linear classifiers, we can further create an ensemble of the most recent $K=10$ epochs, by averaging the softmax outputs.
\end{itemize}

The performance of the three types of classifiers are shown in Figure~\ref{fig-performance-comp}. The classifier ensemble is able to reach the maximal performance of ResNet-18 at an earlier training stage and the performance remains stable. This suggests that in order to reach the final performance of the given model, we can build a classifier ensemble as a proxy instead of train the model until convergence, which greatly reduces the computational cost.

\begin{figure}[h]
\begin{center}
\includegraphics[width=1.0\linewidth]{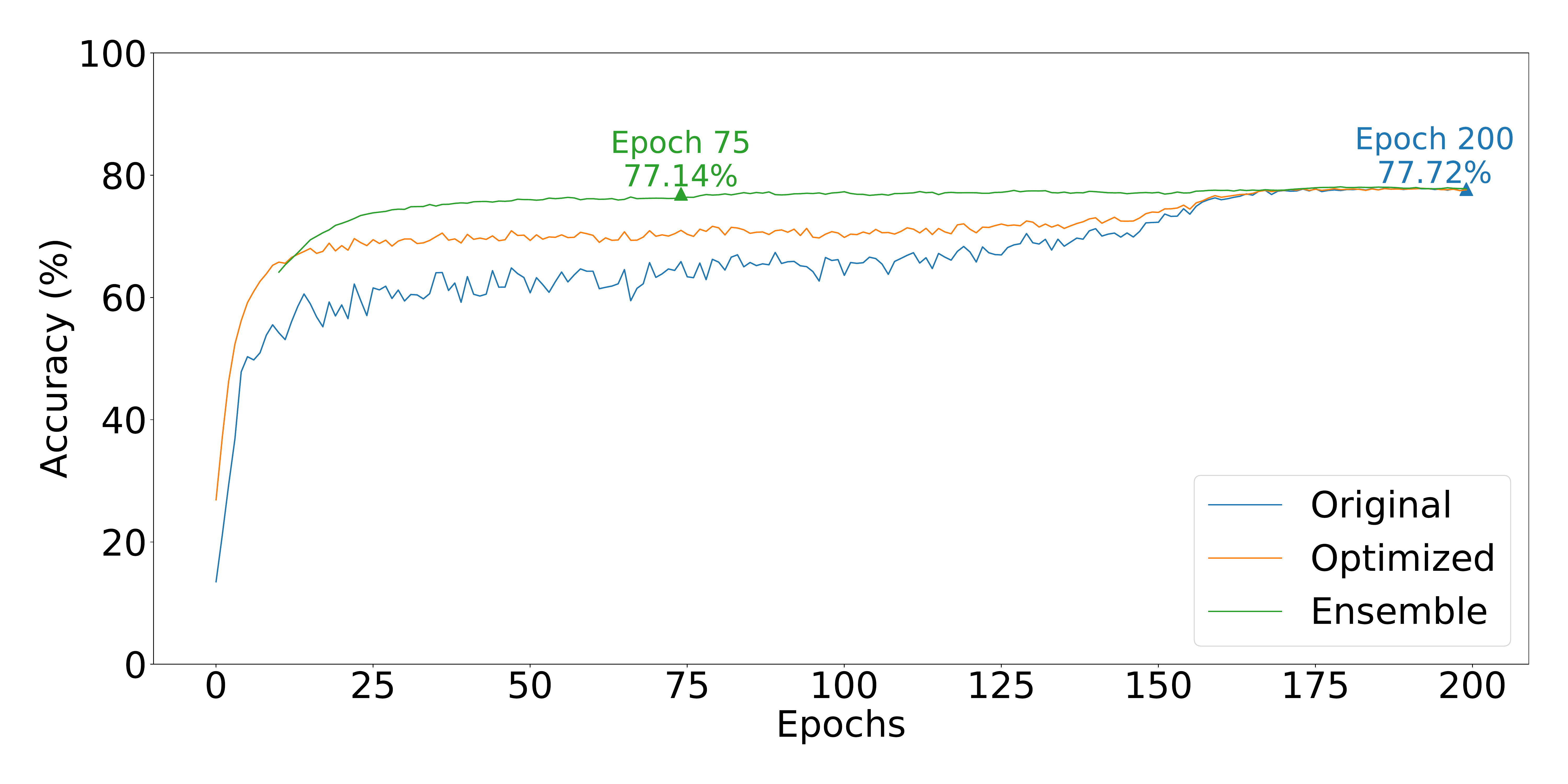}
\end{center}
\caption{Performance of the original classifiers, optimized classifiers, and classifier ensembles in training ResNet-18 on CIFAR-100. In this example, the classifier ensemble already reaches over $77\%$ accuracy at epoch 75, which speeds up the training by about 2.7x compared to full training till convergence.}
\label{fig-performance-comp}
\end{figure}

\begin{figure*}[h]
\centering
\begin{subfigure}{0.3\linewidth}
    \centering
    \includegraphics[width=\linewidth]{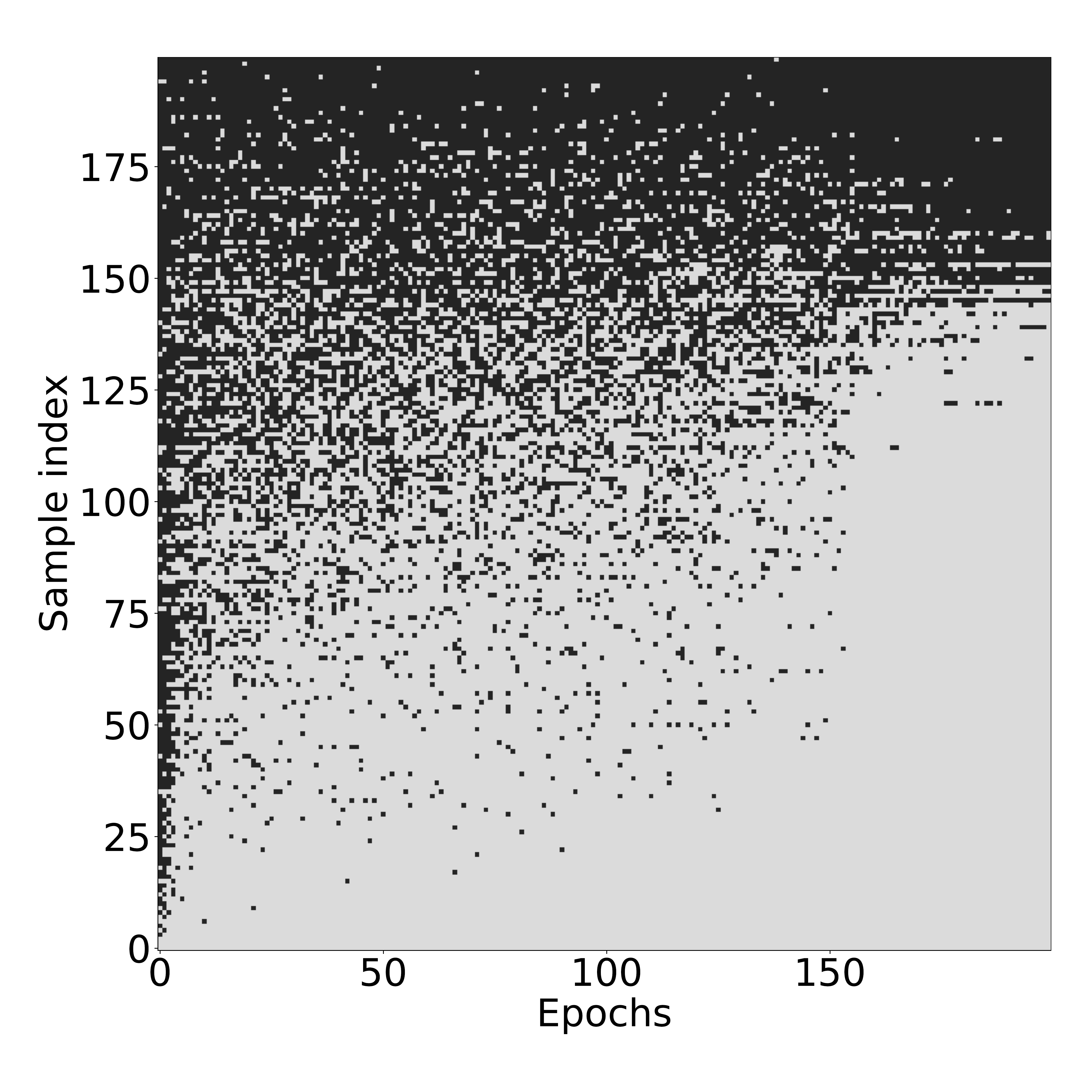}
    \caption{Original}
\end{subfigure}
\hfill
\begin{subfigure}{0.3\linewidth}
    \centering
    \includegraphics[width=\linewidth]{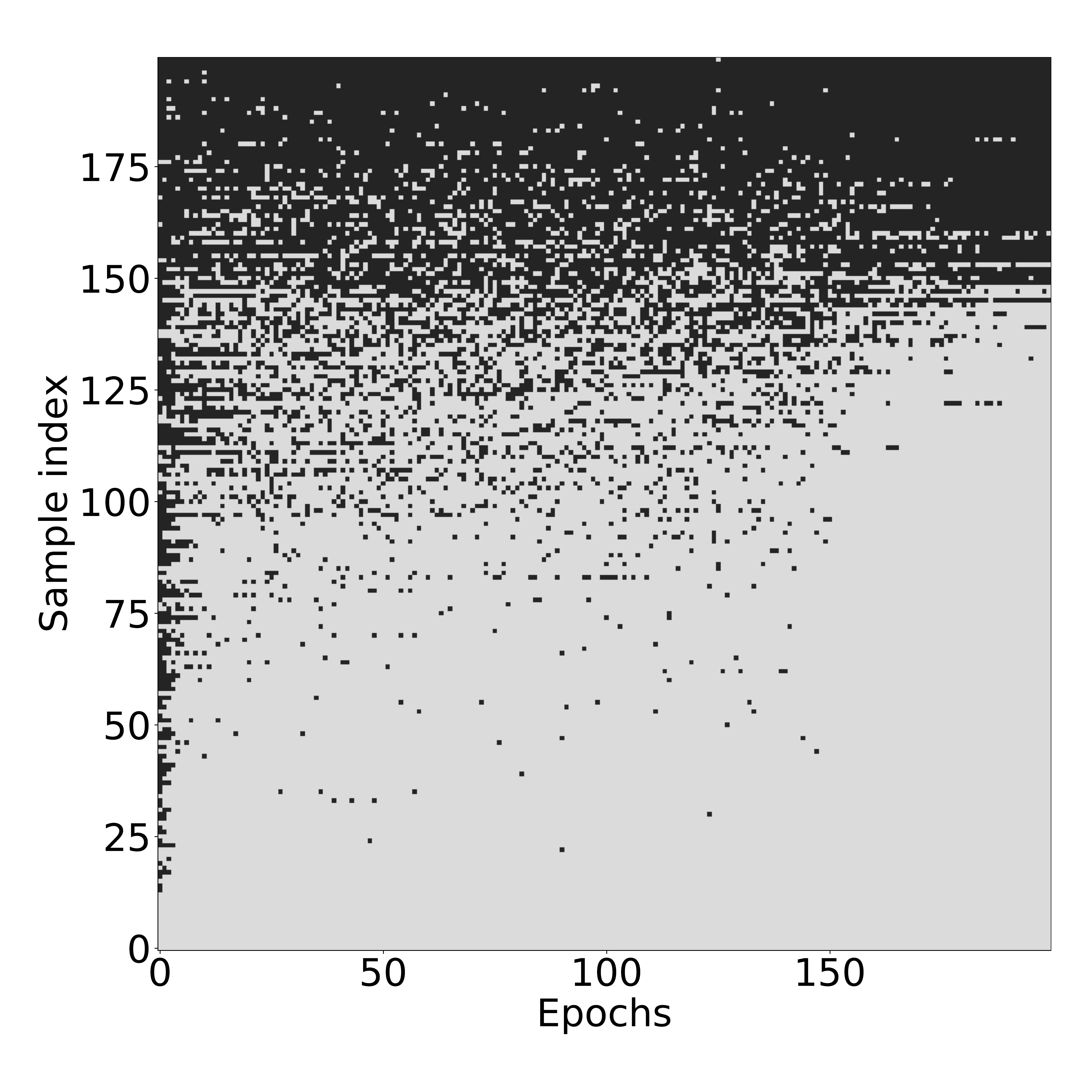}
    \caption{Optimized}
\end{subfigure}
\hfill
\begin{subfigure}{0.3\linewidth}
    \centering
    \includegraphics[width=\linewidth]{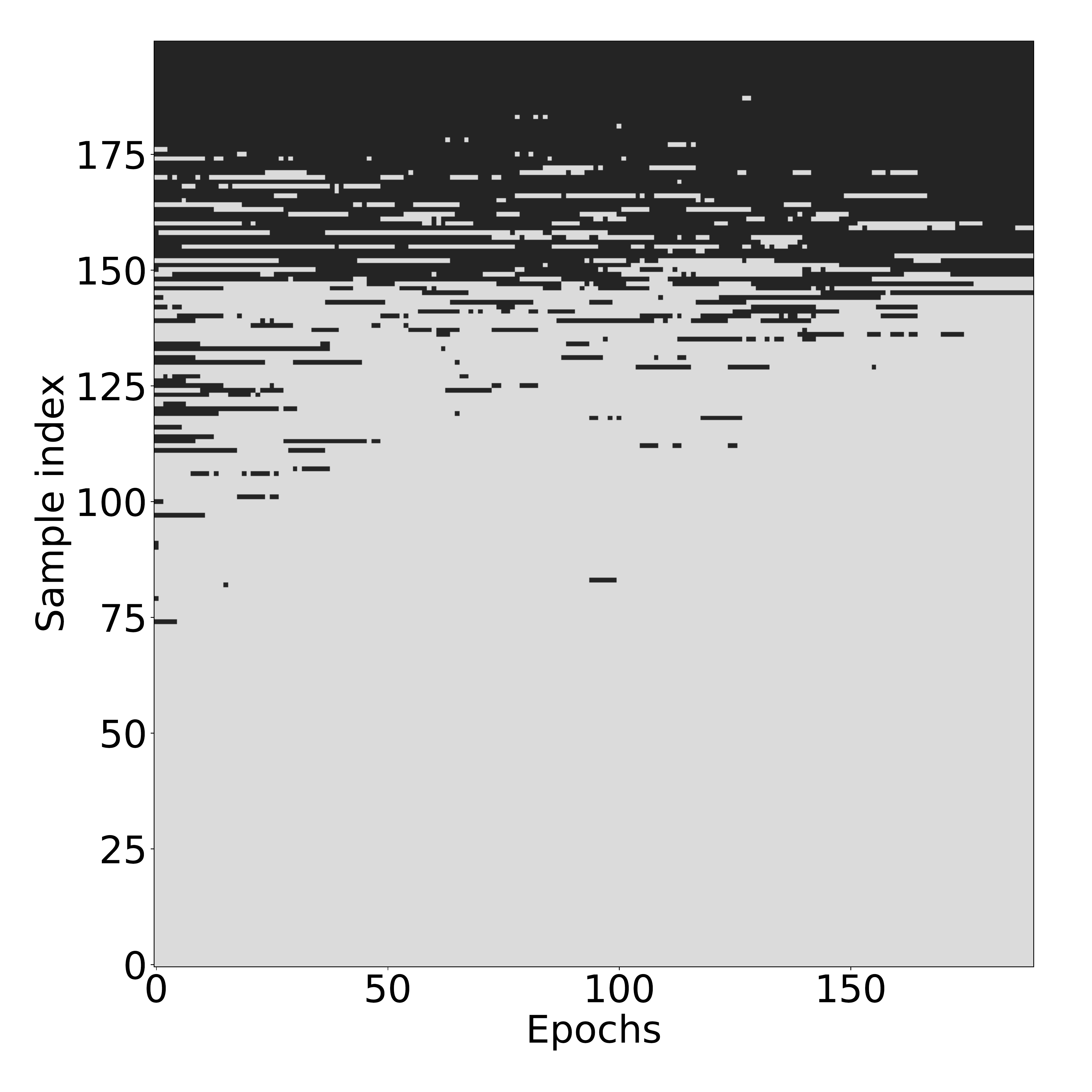}
    \caption{Ensemble}
\end{subfigure}
\caption{Classification results of some sampled validation images during training ResNet-18 on CIFAR-100. A light gray pixel denotes that image is correctly classified by the classifier at the given epoch, and a black pixel denotes a wrong classification. The vertical axis indicates the image indices, sorted by their accuracy for better visualization.}
\label{fig-performance-qual}
\end{figure*}

\begin{figure*}[h]
\centering
\begin{subfigure}{0.33\linewidth}
    \centering
    \includegraphics[width=\linewidth]{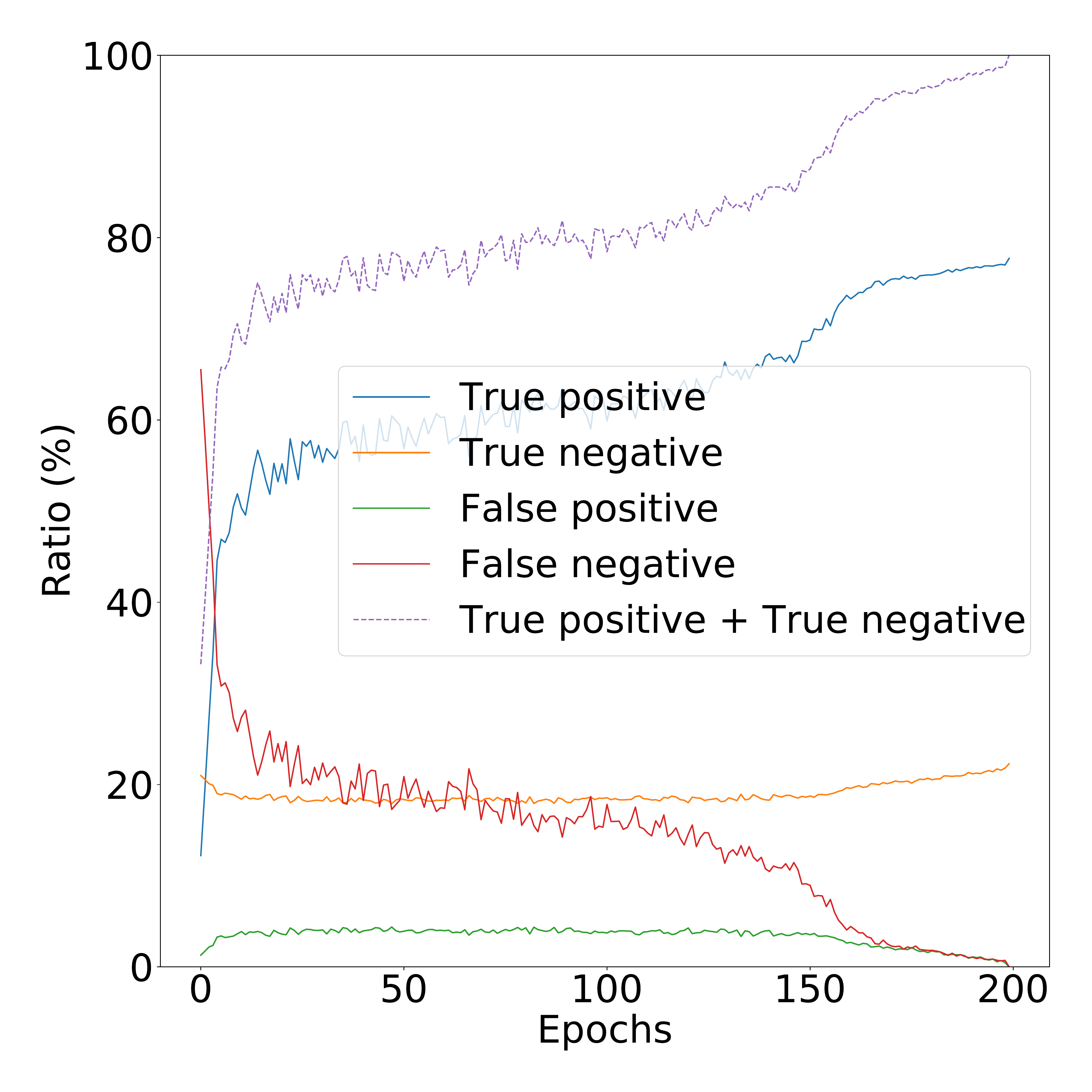}
    \caption{Original}
\end{subfigure}
\hfill
\begin{subfigure}{0.33\linewidth}
    \centering
    \includegraphics[width=\linewidth]{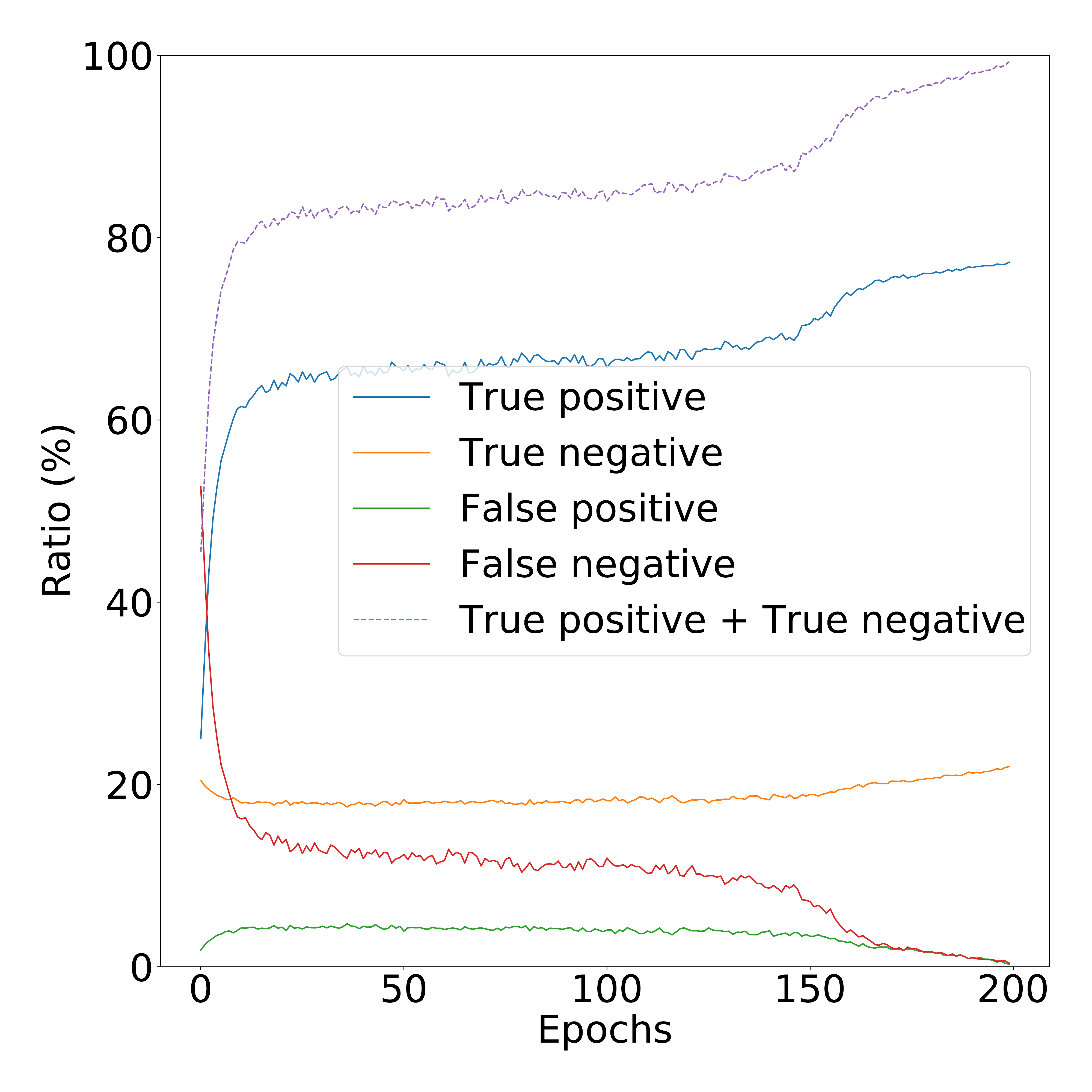}
    \caption{Optimized}
\end{subfigure}
\hfill
\begin{subfigure}{0.33\linewidth}
    \centering
    \includegraphics[width=\linewidth]{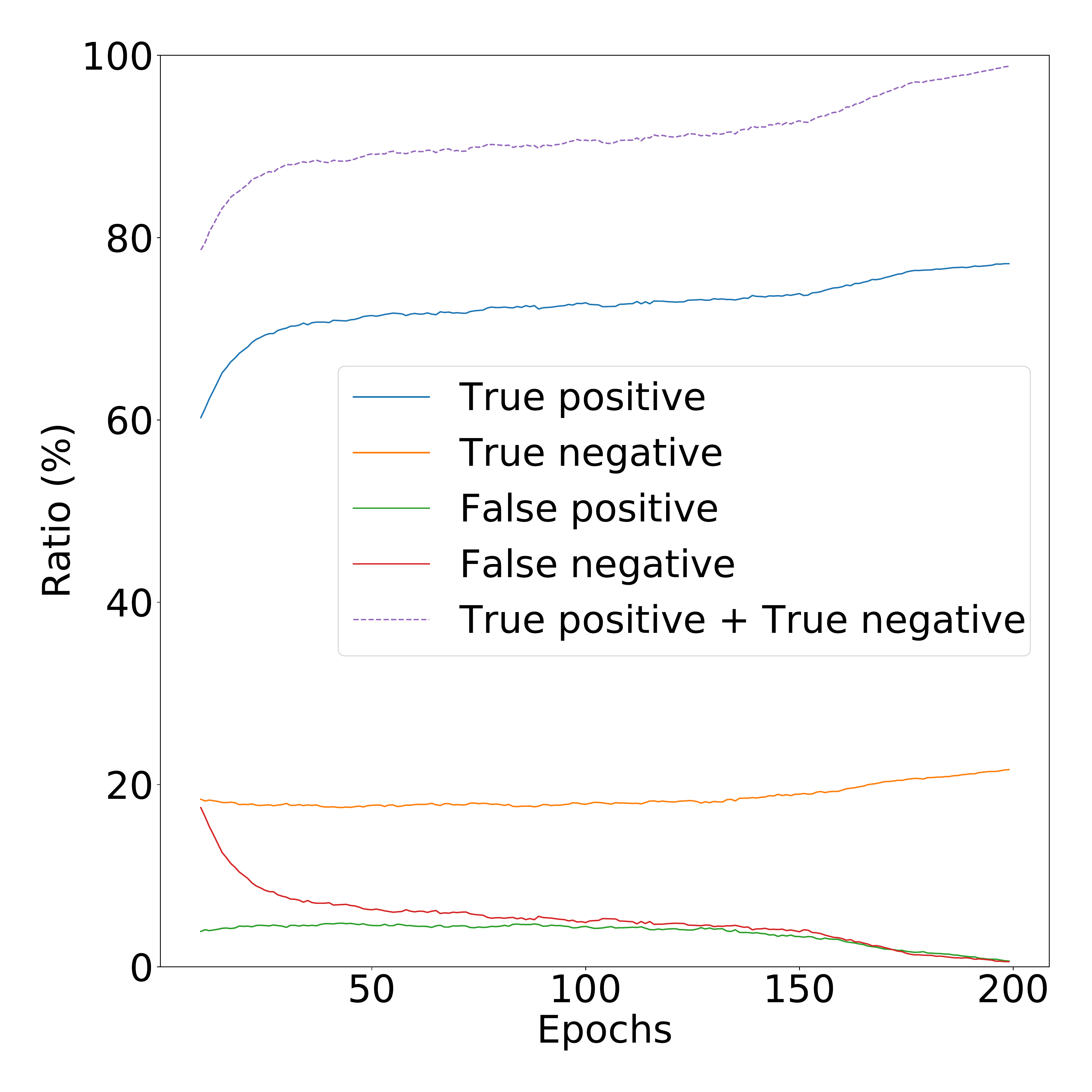}
    \caption{Ensemble}
\end{subfigure}
\caption{Breakdown of classification results of validation images during training ResNet-18 on CIFAR-100. Here we compare the classification results with the final model. True positive: Both the current classifier and the final ResNet-18 can correctly classify. True negative: Both the current classifier and the final ResNet-18 cannot correctly classify. False positive: The current classifier is correct but the final ResNet-18 is wrong. False negative: The current classifier is wrong but the final ResNet-18 is correct.}
\label{fig-performance-quan}
\end{figure*}

We can also take a closer look at the individual validation images. Figure~\ref{fig-performance-qual} shows the qualitative classification results of three types of classifiers by sampling $200$ validation images. Original classification results are rather noisy and unstable, which leads to under-estimated performance in early training stages. Even with the optimized linear classifiers, most validation image features keep moving back and forth between the correct and wrong sides of the optimal linear boundaries. Nevertheless, with classifier ensembles, we are able to obtain stable classification results, whose decisions are more consistent with the fully trained model. Figure~\ref{fig-performance-quan} shows the breakdown of classification results. This further demonstrates that the classifier ensemble can not only reach the maximal performance earlier, but also make similar decisions for most images with the final model. In conclusion, the classifier ensemble can act as an excellent proxy for simulating the behavior of the converged model, thus providing more accurate performance estimation at an early training stage.

\subsubsection{Classifier Ensemble via Feature Histories}

Based on this key observation of classification oscillation and consistency between historical models, we propose to use the classifier ensemble based on the history information of deep features to approximate the final performance of a CNN at an early training stage. Given saved feature histories, we can obtain the optimal linear classifiers for each checkpoint. Then we can test the features of validation images with these classifiers, and decide if the features are correctly positioned most of the time during training. Thus, we can approximate the set of images which will be correctly classified at the end of training, and give a more accurate estimation of the final performance.

The implementation is straight-forward and simple, as summarized in Algorithm~\ref{alg-ours}. In addition to a typical CNN training loop, we only need to explicitly save intermediate features (line 4, 7), optimize the linear classifiers based on saved features (line 9, 10, 11), and ensemble them (line 12, 13) to acquire a performance estimation closer to the final performance that the network can achieve. More details are discussed in {\ifappendix {Appendix~\ref{ap-ensemble}} \else{the supplement}\fi}.

\begin{algorithm}[h]
    \caption{Training loop with our performance estimation}
    \label{alg-ours}
    \KwIn{Initialized network $f(\vx;\vw^{fea}_0,\vw^{cls}_0)$,\newline
    training dataset $(\mX^{train},\mY^{train})$,\newline
    validation dataset $(\mX^{val},\mY^{val})$,\newline
    epoch budget $N$, window size $K$}
    \KwOut{Improved performance estimation $\bar E_N$}
    \For{$i=1$ to $N$}
    {
        Set $\vw^{fea}_i,\vw^{cls}_i=\vw^{fea}_{i-1},\vw^{cls}_{i-1}$\\
        \For{Sampled batch $\mX^{train}_{i,j}, \mY^{train}_{i,j}$ in $\mX^{train},\mY^{train}$}
        {
            Compute and save intermediate features $\mH^{train}_{i,j}=g(\mX^{train}_{i,j};\vw^{fea}_i)$\\
            Compute outputs $\mZ^{train}_{i,j}=h(\mH^{train}_{i,j};\vw^{cls}_i)$\\
            Optimize $\vw^{fea}_i,\vw^{cls}_i$ with one gradient step\\
        }
        Compute and save intermediate features $\mH^{val}_i=g(\mX^{val};\vw^{fea}_i)$\\
        Save checkpoint $\vw^{fea}_i,\vw^{cls}_i$
    }
    \For{$k=N-K+1$ to $N$}
    {
        Initialize $\vv^{cls}_k=\vw^{cls}_k$\\
        Optimize $\vv^{cls}_k$ with $\text{Loss}(h(\mH^{train}_k;\vv^{cls}_k),\mY^{train})$\\
        Compute outputs $\mZ^{val}_k=h(\mH^{val}_k;\vv^{cls}_k)$
    }
    Average predicted distributions $\bar\mZ^{val}_N=\text{Mean}(\mZ^{val}_{N-K+1},\dots,\mZ^{val}_N)$\\
    Evaluate performance $\bar E_N=\text{Eval}(\bar\mZ^{val}_N,\mY^{val})$\\
    \Return{$\bar E_N$}
\end{algorithm}

One concern may be the extra computational overhead of this performance estimation strategy. In fact, compared with a typical training loop, our method usually only requires less than one half epoch time of training the original network, which is almost negligible since we often need to train the network for multiple epochs. The computational overhead of our method has two main parts: 1) Saving the intermediate features. Note the computation of deep features is a part of the original training process, we only need to slightly modify the network implementation, let it return the intermediate results, and save them in the storage after each epoch. 2) Optimization and ensemble of the linear classifiers. As discussed above, the linear classifiers are computationally cheaper and easier to optimize than the feature extraction part. We also have good weight initialization from the saved checkpoints, so a small number of iterations are adequate for convergence. We also observe the memory usage increase is moderate. The image features that we need to save have reduced dimension compared with earlier layers. Therefore, the overhead only takes up a small fraction of the whole training process.

\subsection{Combination with Search Algorithms}
\label{sec-method-comb}

Our method is able to give accurate performance estimation at an early training stage, thus better assess the quality of a configuration with limited computational resources. In order to apply our method in a search task like NAS and HPO, we need to integrate our performance estimation into existing search algorithms for general purpose. In the experiments, we mainly focus on three algorithms: random search, HyperBand~\cite{li2017hyperband}, and BOHB~\cite{falkner2018bohb}.

In random search, we generate a pool of configurations by random sampling from the search space, evaluate each configuration after training for some fixed budget, and pick the best one based on the latest performance. HyperBand is a bandit strategy which adaptively allocates training resources for configurations in the sample pool based on their current performance. BOHB can be considered as a variant of HyperBand, which includes a Bayesian optimization component to better generate the initial configuration pool. These search algorithms do not have assumptions about the search space, making them applicable to most NAS and HPO tasks. When combined with our method, these search algorithm make decisions based on the improved performance estimation instead of the latest performance acquired. In HyperBand and BOHB, this performance metric also determines the configurations which will be allocated with more budgets.

\section{Experiments}
\label{sec-exp}

We first show that our method can efficiently estimate the performance of various CNN architecture families. Then we combine our method with search algorithms and demonstrate its effectiveness on neural architecture search (NAS) and hyperparameter optimization (HPO).

\subsection{Performance Estimation for CNNs}
\label{sec-exp-performance}

We consider the following architecture families: VGG~\cite{simonyan2014very}, ResNet~\cite{he2016deep}, and MobileNetV2~\cite{sandler2018mobilenetv2} and use the CIFAR-10 and CIFAR-100 dataset~\cite{krizhevsky2009learning}. We train each architecture for $200$ epochs (see experimental details in {\ifappendix {Appendix~\ref{ap-cnns}} \else{the supplement}\fi}). We report the test accuracy every $10$ epochs, which is the original performance evaluated at that epoch. We also apply our method for checkpoints during training every $10$ epochs, which gives an improved estimation of the final performance. The results for comparison are shown in Figure~\ref{fig-performance-architecture}. By utilizing the feature histories during network optimization, our method produces stable and accurate estimation of the final performance at an early training stage for all these CNN architectures.

\begin{figure}[h]
\begin{center}
\includegraphics[width=1.0\linewidth]{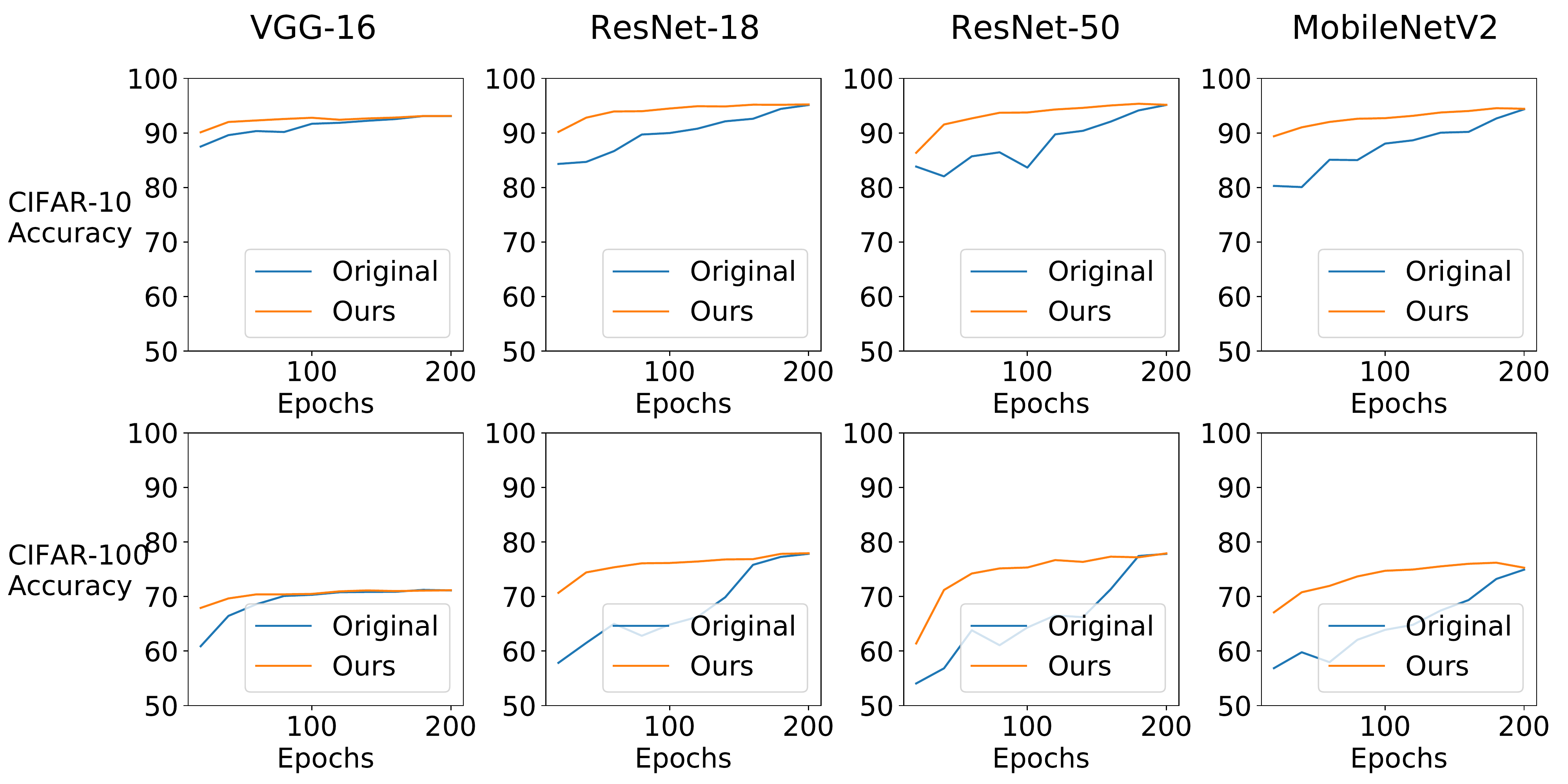}
\end{center}
\caption{Performance estimation for different architectures including VGG-16, ResNet-18, ResNet-50, and MobileNetV2 on CIFAR-10/CIFAR-100 dataset. \textbf{Original} denotes the original test accuracy at each epoch. \textbf{Ours} denotes the performance of our proposed proxy classifier using the feature histories up to the given epoch. Our method can reach an accuracy closer to the final accuracy very early.}
\label{fig-performance-architecture}
\end{figure}

In a resource-limited search setting, one may need to shrink the training budget and adjust the learning rate schedule accordingly, at the cost of reducing the final performance. We also investigate the example of ResNet-18 with fewer training epochs, as shown in Figure~\ref{fig-performance-schedule}. Our method can still accurately approximate the final performance early in this setting.

\begin{figure}[h]
\begin{center}
\includegraphics[width=1.0\linewidth]{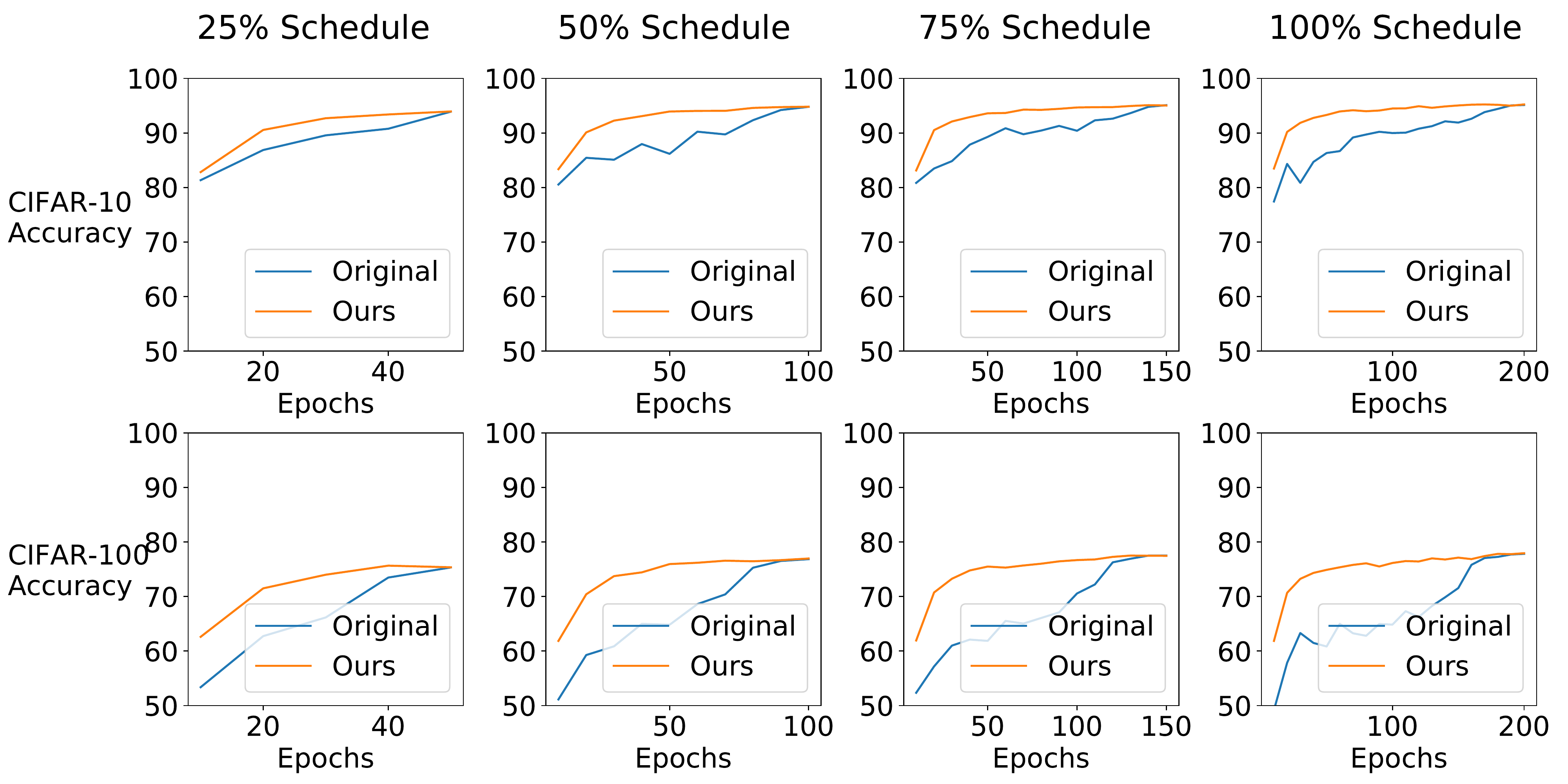}
\end{center}
\caption{Performance estimation for different training budgets from $25\%$ ($50$ epochs) to $100\%$ ($200$ epochs) of training ResNet-18 on CIFAR-10/CIFAR-100 dataset. The linear learning rate schedule is adjusted accordingly.}
\label{fig-performance-schedule}
\end{figure}

\subsection{Neural Architecture Search}

For NAS applications, we conduct experiments in two search spaces: NAS-Bench-201~\cite{dong2020bench} and DARTS~\cite{liu2018darts}.

\subsubsection{NAS-Bench-201}

NAS-Bench-201~\cite{dong2020bench} is a public benchmark for testing NAS algorithms. It defines a search space consisting of $15,625$ architectures, and includes full training log of all the architectures on CIFAR-10, CIFAR-100~\cite{krizhevsky2009learning}, and downsampled ImageNet~\cite{deng2009imagenet}. In the following experiments, we mainly use the information about the ``true performance'' on CIFAR-10, which is defined as the average top-1 test accuracy of three independent runs of training the given architecture for $200$ epochs.

First, we verify that our method is helpful for comparing different architecture configurations. To pick the best architecture among candidates, usually we need to train the architectures for some epochs and evaluate on the validation set to acquire a performance estimate, and the longer we train, the estimate is closer to the true performance. Here we randomly sample $100$ architectures from the search space, train each for $200$ epochs, evaluate both the original models and our proposed proxy classifier every $10$ epochs, and compare their performance with the true performance (see experimental details in {\ifappendix {Appendix~\ref{ap-nas201-ranking}} \else{the supplement}\fi}). Figure~\ref{fig-nas201-error} shows that our method can acquire significantly more accurate estimates for true performance of NAS-Bench-201 architectures.

Besides the absolute numbers, it is important for a NAS algorithm to obtain the relative ranking of architectures that is consistent with the true performance, so that the algorithm can return an architecture among the best ones in the search space. Given the true performance and estimated performance, we can measure the consistency of their relative ranking with the Kendall's $\tau$ rank correlation coefficient~\cite{kendall1938new}. The $\tau$ coefficient ranges in $[-1,1]$, and when the two observations have perfectly matched relative rankings, $\tau$ reaches its maximum value $1$. Figure~\ref{fig-nas201-tau} shows that the performance produced by our method is not only closer to the true performance for each individual architecture, but also keeps better relative ranking for distinguishing different architectures. Using this method, we can find better architectures with limited training budgets.

\begin{figure}[h]
\begin{center}
\includegraphics[width=0.8\linewidth]{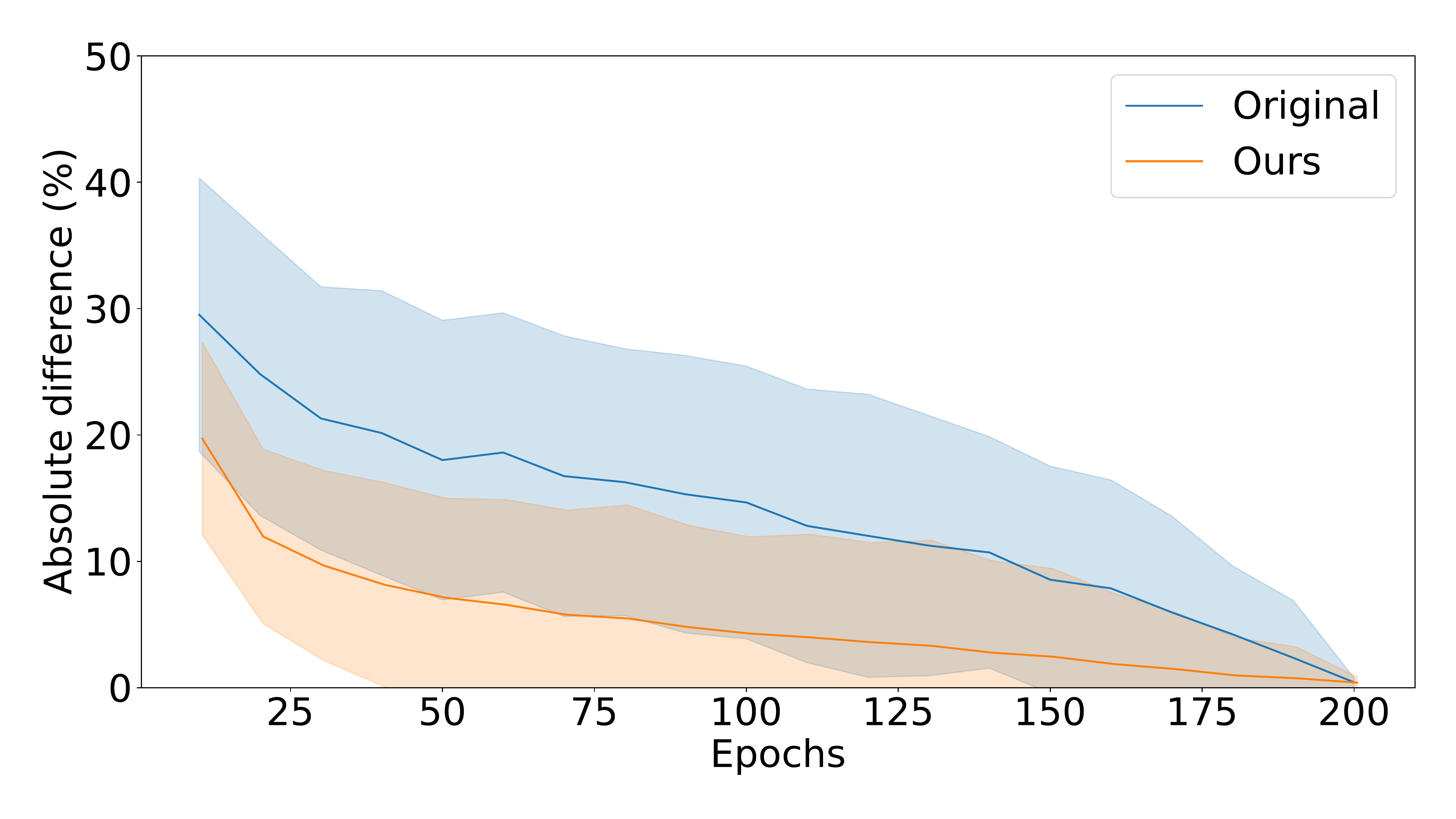}
\end{center}
\caption{Absolute difference between the estimated performance and the true performance. \textbf{Original} denotes the original accuracy metrics at each epoch. \textbf{Ours} denotes the predicted performance by our method using the checkpoints up to the given epochs. For fair comparison, we slightly shift \textbf{Ours} rightwards because of the computational overhead.}
\label{fig-nas201-error}
\end{figure}

\begin{figure}[h]
\begin{center}
\includegraphics[width=0.8\linewidth]{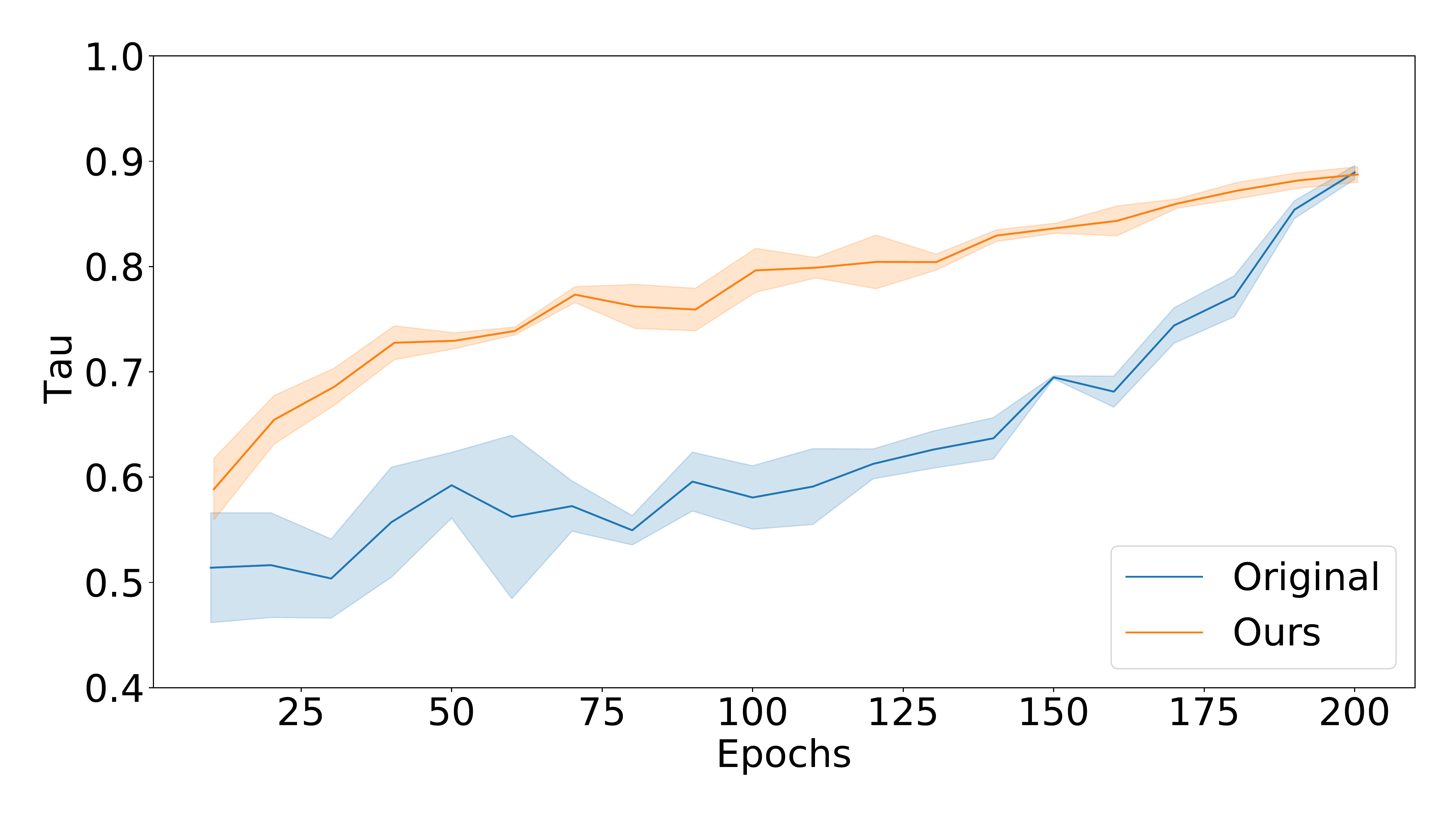}
\end{center}
\caption{Relative ranking measured by Kendall's $\tau$ coefficient between the estimated performance and the true performance.}
\label{fig-nas201-tau}
\end{figure}

\begin{table}[h]
\centering
\caption{Search results for NAS-Bench-201 architectures on CIFAR-10. The global optimal architecture in NAS-Bench-201 search space has accuracy $94.37\%$.}
\label{table-nas201}
\begin{threeparttable}
\begin{tabular}{c|c}
\toprule
Method & Test Accuracy ($\%$) \\
\midrule
Random Search & $93.21\pm0.78$ \\
Random Search + Ours & $93.45\pm0.27$ \\
\midrule
HyperBand & $93.36\pm0.30$ \\
HyperBand + Ours & $\mathbf{93.77\pm0.09}$ \\
\midrule
BOHB & $93.35\pm0.55$ \\
BOHB + Ours & $93.45\pm0.37$ \\
\bottomrule
\end{tabular}
\end{threeparttable}
\end{table}

Next, we compare several search algorithms with and without our performance estimation, as described in Section~\ref{sec-method-comb} (see experimental details in {\ifappendix {Appendix~\ref{ap-nas201-search}} \else{the supplement}\fi}). The search budget is around $2$ hours per run, and we quantify the performance of each search algorithm by the test accuracy of the searched architecture. The results are shown in Table~\ref{table-nas201}. Our method can improve all three search algorithms in terms of the final searched architectures. Especially, the combination of HyperBand and our method can robustly find an architecture which has $<0.7\%$ test accuracy gap from the global optimum.


\begin{table*}[t]
\centering
\caption{Comparison with state-of-the-art NAS methods which uses the DARTS search space on CIFAR-10.}
\label{table-darts}
\begin{threeparttable}
\begin{tabular}{c|cccc}
\toprule
~ & Test Error & Params & Search Cost & Search \\
Method & ($\%$) & (M) & (GPU days / run) & Method \\
\midrule
DARTS (first order) & $3.00\pm0.14$ & $3.3$ & $0.4$ & Gradient \\
DARTS (second order) & $2.76\pm0.09$ & $3.3$ & $1$ & Gradient \\
P-DARTS & $2.50$ & $3.4$ & $0.3$ & Gradient \\
PC-DARTS & $2.57\pm0.07$ & $3.6$ & $0.1$ & Gradient \\
UNAS & $2.53$ & $3.3$ & $4.3$ & Gradient\&RL \\
\midrule
Ours & $2.64\pm0.04$ & $3.4$ & $0.2\times4$\tnote{*} & HyperBand \\
\bottomrule
\end{tabular}
\begin{tablenotes}
\item[*] We use $4$ parallel workers. The wall-clock time is $0.2$ day.
\end{tablenotes}
\end{threeparttable}
\end{table*}

\begin{table*}[t]
\centering
\caption{Search results for scaling factors of ResNet-18 on CIFAR-100.}
\label{table-scaling}
\begin{threeparttable}
\begin{tabular}{c|ccc}
\toprule
Architecture & Test Accuracy ($\%$) & Params (M) & FLOPS (G) \\
\midrule
ResNet-18 & $77.83$ & $11.22$ & $0.56$ \\
\midrule
ResNet-34 & $78.49$ & $21.33$ & $1.16$ \\
$2\times$ ResNet-18 by HB & $78.40\pm0.21$ & $18.35\pm0.42$ & $1.04\pm0.05$ \\
$2\times$ ResNet-18 by HB+Ours & $\mathbf{78.75\pm0.03}$ & $15.83\pm1.75$ & $1.02\pm0.08$ \\
\midrule
$4\times$ ResNet-18 by HB & $79.00\pm0.12$ & $22.10\pm3.36$ & $1.57\pm0.26$ \\
$4\times$ ResNet-18 by HB+Ours & $\mathbf{79.40\pm0.34}$ & $28.31\pm6.23$ & $1.93\pm0.24$ \\
\bottomrule
\end{tabular}
\end{threeparttable}
\end{table*}

\begin{table}[t]
\centering
\caption{Search results for RandAugment hyperparameter optimization on a subset of CIFAR-10.}
\label{table-randaug}
\begin{threeparttable}
\addtolength{\tabcolsep}{-4pt}
\begin{tabular}{c|ccc}
\toprule
~ & \multicolumn{3}{c}{Test Accuracy ($\%$)} \\
Method & VGG-16 & ResNet-18 & MobileNetV2 \\
\midrule
HB & $85.38\pm0.24$ & $85.83\pm0.43$ & $86.53\pm0.23$ \\
HB+Ours & $85.60\pm0.28$ & $86.06\pm0.64$ & $86.62\pm0.48$ \\
\midrule
BOHB & $85.35\pm0.44$ & $86.04\pm1.16$ & $\mathbf{86.92\pm0.46}$ \\
BOHB+Ours & $\mathbf{85.61\pm0.38}$ & $\mathbf{86.53\pm0.64}$ & $86.80\pm0.71$ \\
\bottomrule
\end{tabular}
\addtolength{\tabcolsep}{4pt}
\end{threeparttable}
\end{table}

\begin{table}[t]
\centering
\caption{ImageNet performance of scaled ResNet-18 searched on CIFAR-100.}
\label{table-scaling-imagenet}
\begin{threeparttable}
\begin{tabular}{c|c}
\toprule
Architecture & Top-1 Accuracy ($\%$) \\
\midrule
$2\times$ ResNet-18 by HB & $71.46$ \\
$2\times$ ResNet-18 by HB+Ours & $\mathbf{71.77}$ \\
\midrule
$4\times$ ResNet-18 by HB & $73.01$ \\
$4\times$ ResNet-18 by HB+Ours & $\mathbf{73.14}$ \\
\bottomrule
\end{tabular}
\end{threeparttable}
\end{table}

\subsubsection{DARTS}

Differentiable ARchiTecture Search (DARTS)~\cite{liu2018darts} is a recent NAS algorithm which relaxes the search space to be continuous, achieving search speed orders of magnitude faster than previous methods. Here we directly transfer the method that we use on NAS-Bench-201 to the DARTS search space, to demonstrate our performance estimation strategy is also helpful in another NAS setting.

We use the combination of HyperBand and our method, which performs the best on NAS-Bench-201, and follow the practice of architecture evaluation in DARTS (see experimental details in {\ifappendix {Appendix~\ref{ap-darts}} \else{the supplement}\fi}). The results are summarized in Table~\ref{table-darts}. We also list some more recent DARTS variants which also use the same search space and evaluation configuration, including P-DARTS~\cite{chen2019progressive}, PC-DARTS~\cite{xu2019pc}, and UNAS~\cite{vahdat2020unas}. Our method is able to find better architectures than DARTS with similar search cost. Our result is also close to more recent state-of-the-art NAS methods. It is notable that our method is the only one that does not make use of the gradient-based method, so we can easily parallelize the search process. In fact, we use $4$ parallel workers in the experiment, reducing the wall-clock search time to $0.2$ day per run. We expect to further reduce search time with even more parallel workers.


\subsection{Hyperparameter Optimization}

DARTS and its variants are NAS algorithms customized for their search space design. We are also interested in other tasks where the search space is non-differentiable and thus these NAS algorithms may not apply. Here we investigate two examples of hyperparameter optimization, to show our method is robust and versatile to a wider range of tasks.

\subsubsection{Data Augmentation}

In this task, we consider the problem of optimizing hyperparameters in data augmentation for CNN training. We use the search space defined in RandAugment~\cite{cubuk2020randaugment}, which has only two discrete hyperparameters: $N$ is the number of augmentation transformations to apply, and $M$ is the magnitude for the transformations. The search space is significantly smaller than previous work such as AutoAugment~\cite{cubuk2018autoaugment}, but still computationally expensive for the grid search that \cite{cubuk2020randaugment} does for each CNN architecture and dataset.

To compare HyperBand and BOHB with and without our performance estimation, we search the optimal RandAugment configuration for VGG-16, ResNet-18, and MobileNetV2 on a subset of CIFAR-10 (see experimental details in {\ifappendix {Appendix~\ref{ap-randaugment}} \else{the supplement}\fi}). The results are summarized in Table~\ref{table-randaug}. In most cases, our method consistently improves the baseline search algorithms. This task further demonstrates the effectiveness and versatility of our method.

\subsubsection{Scaling Factors for ResNet}

EfficientNet~\cite{tan2019efficientnet} is a series of CNN architectures with state-of-the-art image classification performance, which is created by compound scaling up the depth, width, and resolution of a small base network. Inspired by EfficientNet, we can consider scaling up the ResNet-18 architecture to build stronger networks by multiplying the number of building blocks and channels at each stage with some scaling factors, under the constraint of FLOPS. We test HyperBand with and without our method, to search the optimal scaling factors for ResNet-18 on CIFAR-100 (see experimental details in {\ifappendix {Appendix~\ref{ap-scaling}} \else{the supplement}\fi}). Table~\ref{table-scaling} shows that our method can help HyperBand to find better scaling factors in this task. We have also transferred the best-performing solutions searched on CIFAR-100 to ImageNet, and the result in Table~\ref{table-scaling-imagenet} further demonstrates the improvement brought by our method.

\section{Conclusion}

We propose a novel efficient model performance estimation method, which effectively use the saved feature histories during optimization to produce accurate approximation of the final performance. Our method is simple to implement and applicable to various tasks in NAS and HPO, and leads to improvement for general search algorithms. For future work, we think it would be interesting to use the performance estimation to guide and accelerate CNN training.

{\small
\bibliographystyle{ieee_fullname}
\bibliography{egbib}
}

\fi

\ifappendix
\clearpage

\ifmain

\else

\title{Supplementary Material for\\Efficient Model Performance Estimation via Feature Histories}
\maketitle

\fi

\appendix
\section{Experimental Details}

\subsection{Classifier Ensemble}
\label{ap-ensemble}

The training procedure of the classifier ensemble has the same following configuration across all experiments. We use the the saved features from the most recent $K=10$ epochs. We first obtain $K$ optimal classifiers by training them on the saved features of images from the training set and their labels. The weights are initialized from the saved network weights of the corresponding epochs. We use the cross entropy loss and SGD optimizer with momentum $0.9$ and weight decay $5\times 10^{-4}$. The learning rate starts from $0.05$ and decays following a linear schedule~\cite{li2019budgeted}. The batch size is $1024$, and the number of epochs is $5$. To build the ensemble, we collect the classification results of the optimized linear classifiers for the validation image features, and output the mean of the softmax-ed probability distribution, which is then used for evaluation and estimation. The whole process introduces little computational cost compared with the original training loop. For example, in NAS-Bench-201, we observe the overhead of our method is typically $0.5$ epoch time for each architecture.

\subsection{Performance Estimation for CNNs}
\label{ap-cnns}

In the first part, we train each CNN on CIFAR-10 and CIFAR-100 for $200$ epochs with batch size $256$. The initial learning rate is set to $0.1$ for ResNet and MobileNetV2, $0.01$ for VGG, and we use a linearly decaying learning rate schedule as suggested by \cite{li2019budgeted}. We use SGD optimizer with momentum $0.9$ and weight decay $5\times 10^{-4}$.

In the second part, we train ResNet-18 on CIFAR-100 for $50,100,150,200$ epochs, corresponding to $25\%,50\%,75\%,100\%$ budgets. The learning rate still follows the linear schedule, which drops to zero at the end of each budget. Other settings are the same as the first part.

\subsection{NAS-Bench-201}
\subsubsection{Performance Error and Relative Ranking}
\label{ap-nas201-ranking}

In each run of this experiment, we first randomly sample $100$ architectures from NAS-Bench-201~\cite{dong2020bench}, and train each for $200$ epochs with batch size $256$. The initial learning rate is set to $0.1$ and we use the linear learning rate schedule. We use SGD optimizer with momentum $0.9$ and weight decay $5\times 10^{-4}$. We split the original CIFAR-10 training set into two subsets: $40000$ images for training and $10000$ images for validation.

For the performance error, we compute absolute difference between the current performance and the true performance, and for the relative ranking, we compute the Kendall's $\tau$ coefficient~\cite{kendall1938new} between the current performance and the true performance. The experiment is repeated $3$ times.

\subsubsection{Architecture Search}
\label{ap-nas201-search}

For random search, we randomly sample $64$ architectures, and train each for $128$ epochs with batch size $256$. For HyperBand and BOHB, the training budget for each architecture ranges in $[1, 128]$ epochs, and the factor for increasing training budget and shrinking sample pool is set as $\eta=2$. We use $4$ parallel workers, each using an NVIDIA GeForce RTX 2080 Ti GPU. The total search time is about the same for the three algorithms, around $2$ hours. During search, the initial learning rate is set to $0.1$ and we use the linear learning rate schedule with the maximal budget set to $128$. We use SGD optimizer with momentum $0.9$ and weight decay $5\times 10^{-4}$. we use a subset from the original CIFAR-10 training set with $8000$ images for training, and $2000$ images for validation. After search, we calculate the regret as the difference in the true performance between the architecture given by each algorithm and the global optimal architecture, provided by the benchmark. Each experiment is repeated $5$ times with different random seeds.

\subsection{DARTS}
\label{ap-darts}

During search, the search space relaxation is not applicable in our method. We still randomly sample from the discrete search space defined in DARTS~\cite{liu2018darts}, and select the best architectures using HyperBand and our performance estimation. We decompose the process of sampling architecture into sampling the operations and connections for each node in the cells from the uniform distribution of valid choices. At search time, we set the batch size to $96$, initial number of channels to $32$, number of cells to $12$. We also use Cutout~\cite{devries2017improved}, path dropout, and auxiliary towers. The initial learning rate is $0.025$ and we use the linear learning rate schedule with the maximal budget set to $128$. We use SGD optimizer with momentum $0.9$ and weight decay $3\times 10^{-4}$. Other search settings are the same as Appendix~\ref{ap-nas201-search}.

For architecture evaluation, we closely follow the setup of DARTS: We repeat the search process for $4$ times and pick the best architecture for fully train and evaluation. An enlarged architecture with $36$ initial channels is trained for $600$ epochs with batch size $96$, using Cutout, path dropout, and auxiliary towers. The only difference in architecture evaluation is we increase the number of cells from $20$ to $22$. The reason is that our searched cells have more parameter-free operations like skip connections and pooling layers compared with DARTS, so we increase the number of cells to match the number of parameters in DARTS.

\subsection{Data Augmentation}
\label{ap-randaugment}

To better show the influence of data augmentation, we only use a small subset of the original CIFAR-10 training set with $8000$ images for training and $2000$ images for validation. Besides RandAugment~\cite{cubuk2020randaugment}, default data augmentation for training images also includes random flips, pad-and-crop and Cutout, following RandAugment. Other search settings are the same as Appendix~\ref{ap-nas201-search}. Each experiment is repeated $3$ times with different random seeds.

\subsection{Scaling Factors for ResNet}
\label{ap-scaling}

We use ResNet-18~\cite{he2016deep} as our base model. The $\{C_2, C_3, C_4, C_5\}$ stages of ResNet-18 have $\{2, 2, 2, 2\}$ basic building blocks and $\{64, 128, 256, 512\}$ channels respectively. We search for $8$ hyper-parameters $\{d_2,d_3,d_4,d_5,w_2,w_3,w_4,w_5\}$ for scaling up the depth and width of ResNet-18. The scaled network has $\{d_2\times 2, d_3\times 2, d_4\times 2, d_5\times 2\}$ blocks and $\{w_2\times 64, w_3\times 128, w_4\times 256, w_5\times 512\}$ channels for each stage. Meanwhile, we constrain the size of the scaled network by its FLOPS. For CIFAR-100, ResNet-18 has $0.56$ GFLOPS, so $2\times$ ResNet-18 should have no larger than $1.12$ GFLOPS, and $4\times$ ResNet-18 should have no larger than $2.24$ GFLOPS. Other search settings are the same as Appendix~\ref{ap-nas201-search}. Each experiment is repeated $3$ times with different random seeds.

\ifmain

\else

{\small
\bibliographystyle{ieee_fullname}
\bibliography{egbib}
}

\fi

\fi

\end{document}

%% file: math_commands.tex
\usepackage{amsmath,amsfonts,bm}

\def\eqref#1{equation~\ref{#1}}

\def\1{\bm{1}}

\def\vv{{\bm{v}}}
\def\vw{{\bm{w}}}
\def\vx{{\bm{x}}}

\def\mH{{\bm{H}}}

\def\mX{{\bm{X}}}
\def\mY{{\bm{Y}}}
\def\mZ{{\bm{Z}}}

\DeclareMathAlphabet{\mathsfit}{\encodingdefault}{\sfdefault}{m}{sl}
\SetMathAlphabet{\mathsfit}{bold}{\encodingdefault}{\sfdefault}{bx}{n}

\def\sX{{\mathbb{X}}}

\newcommand{\R}{\mathbb{R}}

